\documentclass[conference]{IEEEtran}
\IEEEoverridecommandlockouts

\usepackage{amsmath,amssymb,amsfonts}
\usepackage{bbm}
\usepackage{algorithmic}
\usepackage{graphicx}
\usepackage{textcomp}
\usepackage{xcolor}
\usepackage{todonotes}
\usepackage{lipsum}
\usepackage{biblatex}
\usepackage{standalone}
\usepackage{subcaption}
\usepackage{soul}

\begin{document}

\title{Task-specific Subnetwork Discovery in Reinforcement Learning for Autonomous Underwater Navigation}

\author{
\IEEEauthorblockN{
Yi-Ling Liu\IEEEauthorrefmark{1},
Melvin Laux\IEEEauthorrefmark{1}\IEEEauthorrefmark{2},
Mariela De Lucas Alvarez\IEEEauthorrefmark{1},
Frank Kirchner\IEEEauthorrefmark{1}\IEEEauthorrefmark{2},
Rebecca Adam\IEEEauthorrefmark{1}
}
\IEEEauthorblockA{\IEEEauthorrefmark{1}
Robotics Innovation Center,
German Research Center for Artificial Intelligence,
Bremen, Germany}
\IEEEauthorblockA{\IEEEauthorrefmark{2}
Robotics Research Group,
University of Bremen,
Bremen, Germany}
Correspondence: yi-ling.liu@dfki.de
}

\maketitle

\begin{abstract}
Autonomous underwater vehicles are required to perform multiple tasks adaptively and in an explainable manner under dynamic, uncertain conditions and limited sensing, challenges that classical controllers struggle to address. This demands robust, generalizable, and inherently interpretable control policies for reliable long-term monitoring.
Reinforcement learning, particularly multi-task RL, overcomes these limitations by leveraging shared representations to enable efficient adaptation across tasks and environments.
However, while such policies show promising results in simulation and controlled experiments, they yet remain opaque and offer limited insight into the agent’s internal decision-making, creating gaps in transparency, trust, and safety that hinder real-world deployment. The internal policy structure and task-specific specialization remain poorly understood.
To address these gaps, we analyze the internal structure of a pretrained multi-task reinforcement learning network in the HoloOcean simulator for underwater navigation by identifying and comparing task-specific subnetworks responsible for navigating toward different species.
We find that in a contextual multi-task reinforcement learning setting with related tasks, the network uses only about 1.5\% of its weights to differentiate between tasks. Of these, approximately 85\% connect the context-variable nodes in the input layer to the next hidden layer, highlighting the importance of context variables in such settings.
Our approach provides insights into shared and specialized network components, useful for efficient model editing, transfer learning, and continual learning for underwater monitoring through a contextual multi-task reinforcement learning method.
\end{abstract}

\section{Introduction}
Reinforcement learning (RL) offers the potential for autonomous underwater vehicles to adaptively navigate complex environments, where traditional model-based control methods struggle due to disturbances, partial observability and changing environmental conditions \cite{Christensen2022}.

Although RL for robotic navigation has advanced recently, trust and safe deployment in real-world missions are limited by the lack of interpretability and incomplete understanding of internal decision-making processes, which is particularly critical for autonomous underwater vehicle (AUV) control models. For long-term underwater monitoring tasks, where failures could result in mission loss or environmental risk, addressing this challenge is essential.

Concurrently, multi-task reinforcement learning (MTRL) is theoretically expected to exploit shared knowledge across related tasks \cite{DEramo2019}, thereby enhancing generalization, improving data efficiency, and increasing robustness to variable conditions, which are highly advantageous in the uncertain and non-stationary dynamics of underwater environments. Contextual MTRL further allows task specification through context variables or the incorporation of environment-specific information, such as currents, which are known to influence the underlying Markov decision processes (MDPs).

\begin{figure}
    \centering
    \includegraphics[width=0.98\linewidth]{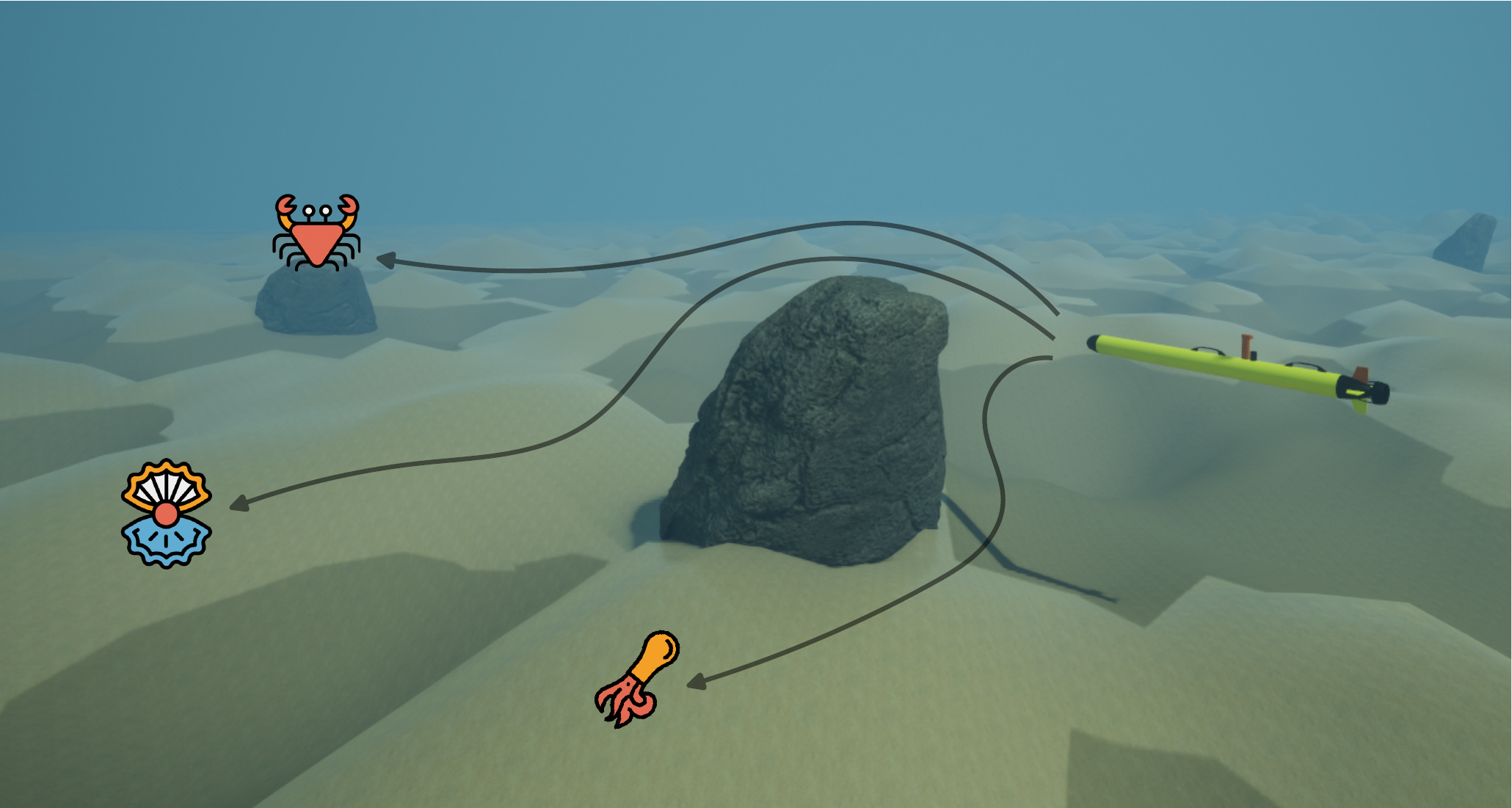}
    \caption{The navigation task is simulated in HoloOcean. For the specified species in each task, the AUV should navigate to find the crab, the shell or the octopus.}
    \label{fig:holoocean}
\end{figure}

\subsection{Related Work}
Despite theoretical and empirical evidence suggesting that MTRL benefits from shared knowledge across tasks \cite{DEramo2019}, there has been limited direct investigation into the internal structure of these networks. This gap is particularly notable from a mechanistic interpretability \cite{Bereska2024} perspective. Even in the single-task reinforcement learning setting, explainability research remains relatively underexplored compared to other areas of machine learning \cite{Milani2024}.

With the recent rise of mechanistic interpretability, driven by advances in large language models, there has been growing interest in analyzing the internal mechanisms of reinforcement learning systems. However, existing work remains limited to specialized settings and architectures. Prior studies primarily investigate toy problems, such as goal misgeneralization in maze-solving tasks \cite{Trim2024}, planning in simplified box-pushing environments \cite{Bush2025}, and memory usage in RNN-based world models in the MiniGrid Switching Memory environment \cite{Sobotka2025}. To the best of our knowledge, no prior work has examined the internal structure or mechanisms of contextual MTRL networks.

\begin{figure}
    \centering
    \resizebox{1.0\linewidth}{!}{
        \input{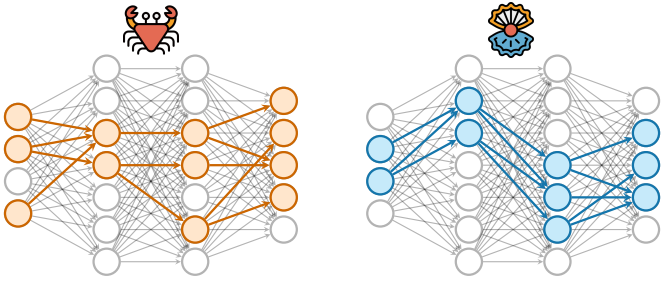}
    }
    \caption{Task‑specific subnetworks specialized for navigation to the crab and to the shell, respectively.}
    \label{fig:subnetworks}
\end{figure}

\subsection{Research Gap}
Motivated by these considerations, we investigate a pretrained MTRL network presented in \cite{Melvin2026} to examine whether its parameters are effectively shared across tasks. In this paper, we further aim to identify task-specific subnetworks within the trained MTRL network that are sufficient for individual navigation tasks, such as reaching distinct target locations or monitor specific marine species. Characterizing these subnetworks can reveal failure modes, enable more efficient inference, and facilitate knowledge transfer across tasks. Ultimately, this contributes to the development of safer, more reliable, and computationally efficient autonomous navigation policies for underwater applications.

In summary, we address following research gaps:
\begin{itemize}
    \item AUV control model lacks interpretability, which is needed for safe real-world deployment.
    \item It remains an open question whether contextual MTRL encodes common knowledge across related tasks or acquires it independently for each task.
    \item It is unclear whether contextual MTRL relies on context variables or other parameters to differentiate tasks.
\end{itemize}

\subsection{Research Questions}
To address these gaps, We seek to answer the following research questions through this study:
\begin{itemize}
    \item RQ1: How are parameters shared in contextual MTRL for AUV control?
    \item RQ2: How are task-specific context variables associated with their respective tasks in contextual MTRL?
    \label{research_questions}
\end{itemize}

To address these research questions, we employ a pretrained Double DQN \cite{Hasselt2015} value network for underwater navigation in \cite{Melvin2026}. For each task, the network is pruned to obtain a task-specific subnetwork, and the overlap between these subnetworks is analyzed to understand shared and task-specific components across different navigation tasks. Building on prior work in neural network modularity and subnetwork discovery, we hypothesize that shared components encode critical knowledge for generalization, while specialized components encode specific strategies. Initial experiments were conducted in MiniGrid \cite{MinigridMiniworld23} to establish the methodological feasibility and then extended to underwater navigation tasks using the HoloOcean simulator \cite{Potokar2024}. By comparing subnetworks across tasks, we identify task-specific structures that differentiate tasks in MTRL, while also uncovering important shared structures that can guide future transfer learning and the reuse of learned knowledge.

\subsection{Novel Contribution}
In summary, the main contributions of this work are:
\begin{itemize}
    \item Demonstrating that MTRL utilizes a large portion of network weights for shared knowledge across tasks, indicating effective knowledge sharing as intended.
    \item Identifying that only a relatively small portion of weights are task-specific, highlighting the minimal task-dependent specialization required for individual objectives.
    \item Revealing the importance of context variables in MTRL, which enable the network to differentiate between related tasks effectively.  
\end{itemize}

\subsection{Notational Conventions}
In this work we use bold lower case letters to denote vectors. Furthermore, $\odot$ denotes the element-wise multiplication (Hadamard Product), and we denote binary sets by $\mathbb{B}=\{0,1\}$.
Finally  $\mathbbm{1}_{\text{condition}}$ denotes the indicator function to represent binary decisions

\begin{equation}
\mathbbm{1}_{\text{condition}} =
\begin{cases}
1, & \text{if the condition is true},\\
0, & \text{otherwise}.
\end{cases}
\end{equation}

It converts a logical condition into a binary value which we use to denote whether a given element satisfies a specific criterion.

\section{Related Work}
Recent research has shown that large neural networks often contain smaller functional subnetworks capable of performing specific tasks, suggesting that complex models may internally organize into modular computational structures.

\subsection{Lottery Ticket Hypothesis}
The Lottery Ticket Hypothesis \cite{Frankle2019} demonstrated that dense neural networks contain sparse subnetworks that can be trained in isolation from their initial weights while achieving comparable performance to the original model. Subsequent work provided theoretical insights into this phenomenon, showing that effective subnetworks can emerge through pruning processes and may explain why overparameterized networks generalize well \cite{Malach2020}.

Later studies further suggested that multiple performant subnetworks may coexist within a single model, supporting the idea that networks may encode different capabilities in partially independent structures \cite{Diffenderfer2021}.

\subsection{Emergent Modularity}
Complementary studies attempt to identify modules responsible for specific subtasks in  trained full networks through weight pruning. Csordás et al. (2021) \cite{Csordas2021} proposed learning differentiable binary masks over weights using a Gumbel-Sigmoid relaxation \cite{Jang2017}, allowing the discovery of functional modules within trained networks. However, most subnetworks were not reused across tasks, indicating a lack of compositionality. Later work \cite{Lepori2023} showed that language and vision models often naturally exhibit structural compositionality without explicit guidance, decomposing tasks into modular subnetworks, suggesting that neural networks can learn compositionality.

These studies provide mixed evidence regarding functional modularity in neural networks, motivating further investigation into whether task-specific subnetworks also arise within MTRL networks.

\subsection{Circuit Discovery}
A closely related line of research is circuit discovery within the emerging field of mechanistic interpretability \cite{Bereska2024}. Circuits refer to sparse computational subgraphs of a model that encode specific aspects of its behavior. Circuit discovery aims to find which parts of a neural network are causally responsible for a specific behavior. Most existing work focuses on pretrained large language models \cite{Conmy2023}.

However, the proposed methods are primarily tailored to transformer architectures, typically operating by pruning connections between high-level components such as entire MLP blocks or attention heads. As a result, they are not directly applicable to simpler or fundamentally different network architectures.

In RL, this research direction remains relatively underexplored. Sobotka et al. (2025) \cite{Sobotka2025} investigate the internal representations of recurrent RL agents, identifying the minimal subnetwork responsible for memory-based behavior in DreamerV3 \cite{Hafner2024}, a model-based RL agent with a GRU-based recurrent backbone \cite{Cho2014}.

Nevertheless, research on the internal mechanisms of RL agents remains limited, with most existing work focusing on specific architectures. Canonical model-free algorithms (e.g., DQN, PPO, SAC) remain poorly understood, and further work is needed to clarify their internal computation and decision-making. In addition, MTRL, despite its potential for shared representation learning, is still underexplored from the perspective of interpretability and needs further investigation.

\section{preliminaries}

\subsection{Contextual Markov Decision Process and Multi-Task Reinforcement Learning}
A contextual MDP \cite{Hallak2015, Modi2018} is defined as a tuple $(\mathcal{C}, \mathcal{S}, \mathcal{A}, \mathcal{M})$, where $\mathcal{C}$ denotes a context space, $\mathcal{S}$ and $\mathcal{A}$ are shared state and action spaces, and $\mathcal{M}$ is a mapping from contexts to tasks. For each context $c \in \mathcal{C}$, this mapping specifies a task $\mathcal{T}_c = (\mathcal{S}, \mathcal{A}, P_c, R_c)$, where the transition dynamics $P_c$ and reward function $R_c$ may vary across contexts.

In MTRL, the objective is to learn a single model that performs well across a distribution of such tasks \cite{Varghese2020}. This is commonly achieved by conditioning the policy or value function on the context, e.g., learning a Q-function $Q(s,a,c)$ that captures task-dependent behavior. Training proceeds by sampling trajectories from multiple contexts and optimizing a joint objective over the task distribution, enabling the agent to share knowledge and generalize across related tasks.

\section{Methodology}
We formulate underwater navigation as a contextual MTRL problem as in \cite{Melvin2026}. The tasks are encoded as a one-hot vector, with each variable representing a navigation goal toward a specific species. Unlike traditional RL, which learns each task independently, MTRL trains a single agent to solve multiple tasks simultaneously by leveraging shared knowledge. In theory, this enables more efficient learning and better generalization through shared representations.
To answer our research questions in Section \ref{research_questions}, we identify task-associated subnetworks and analyze their shared and task-specific structures.

\subsection{Identifying Task-specific Subnetworks}
We identify task-specific subnetworks in a pretrained Q‑network at the level of individual weights. To isolate subnetworks defined by the network parameters, we follow subnetwork discovery approaches similar to \cite{Sobotka2025}, \cite{Csordas2021}, and \cite{Bayazit2024}, where binary masks are learned over pretrained, frozen weights. For the ease of subsequent notation, we conveniently flatten the complete Q-network's concatenated weight matrices including weights across all layers to a single vector formulation $\mathbf{w}=[w_1,\hdots,w_N] \in \mathbb{R}^{N\times 1}$, where $N$ is the total number of network weights. Furthermore, we define a corresponding binary mask vector $\mathbf{m}=[m_1,\hdots,m_N]\in\mathbb{B}^{N\times 1}$ that is multiplied element-wise to the weights to produce the final masked parameters $\mathbf{w}^{\prime} \in \mathbb{R}^{N\times 1}$:
\begin{equation}
\mathbf{w}^{\prime} = \mathbf{w} \odot \mathbf{m}.
\label{eq:get_final_weights}
\end{equation}
Here, each mask element $m_i\in\{1,\hdots,N\}$ in (\ref{eq:get_final_weights}) is given by
\begin{align}
m_i = \mathbbm{1}_{p_i > 0.5}=
\begin{cases}
1, & p_i > 0.5,\\
0, & \text{else}.
\end{cases}
.
\label{eq:binary_mask_thresholding}
\end{align}

The corresponding masking probability $p_i$ in (\ref{eq:binary_mask_thresholding}) depends on learnable parameters logit $l_i \in \mathbb{R}$ and is determined by the sigmoid function such that:
\begin{equation}
p_i = \sigma(l_i) = \frac{1}{1 + e^{-l_i}}.
\end{equation}
The mask logits $\mathbf{l}$ are initialized with small Gaussian noise and constitute the only trainable parameters. Note that, during training the formulation $p_i=\sigma(l_i)$ and $l_i$ serve the purpose of providing a differentiable parametrization for gradient-based optimization in the back-propagation step.
Importantly, masks are learned exclusively for the weights, while all bias parameters are kept fixed and remain unmasked.

The mask parameters are optimized by sampling states directly from the RL state space with the corresponding task context.
This enables the extraction of a sparse subnetwork that preserves the Q-value estimation of the original network on the target task, yielding a compact set of task-relevant connections, as illustrated in Fig.~\ref{fig:subnetworks}.

\subsection{Mask Training Pipeline}
To enable gradient-based optimization of binary masks, differentiable mask training relies on a continuous relaxation of the discrete masking operation. This is necessary because binary variables are non-differentiable and cannot be directly optimized with gradient descent. The method therefore optimizes continuous mask parameters and bridges the gap between discrete forward computation and continuous optimization using the straight-through estimator \cite{Bengio2013}. This applies hard binary decisions in the forward pass while using a differentiable surrogate during backpropagation, enabling learning of discrete masks within standard gradient-based training.

\paragraph{Mask Training}
The sigmoid output $\sigma(l_i)$ serves as a differentiable relaxation of the binary mask, allowing gradients to flow through the masking process. As a result, optimization is performed over the unconstrained logits $l_i$ rather than probabilities, which ensures stable gradient-based updates while maintaining $p_i \in [0,1]$.

During training, masked weights are computed using the straight-through estimator:
\begin{equation}
\tilde{w}_i = w_i \Big( [m_i - \sigma(l_i)]_{\mathrm{stop}} + \sigma(l_i) \Big).
\end{equation}

Here, $w_i$ denotes pretrained network weights, which remain fixed throughout optimization. The stop-gradient operator indicates that the term inside $[\cdot]_{\mathrm{stop}}$ is used only in the forward pass and does not contribute to gradients during backpropagation. This separates the discrete masking decision from the gradient-based optimization path, which instead flows through the differentiable surrogate $\sigma(l_i)$. As a result, only the logits $l_i$ are updated, and the binary mask is learned indirectly via its continuous parameterization.

\paragraph{Objective}
The mask parameters are learned by employing a combined loss $L_{\text{final}}$ balancing task performance and sparsity:
\begin{align}
L_{\text{final}} = L_{\text{q-value}} + \lambda L_{\text{sparsity}},
\end{align}
where $\lambda$ controls the sparsity penalty.

The Q-value loss enforces consistency between the original and masked networks:
\begin{equation}
L_{\text{q-value}} = \mathbb{E}\!\left[\left(Q(s,a) - Q_{\text{masked}}(s,a)\right)^2\right].
\end{equation}

Sparsity is encouraged via an $\ell_1$ penalty on the mask probabilities:
\begin{equation}
L_{\text{sparsity}} = \sum_{i=1}^{N} \sigma(l_i).
\end{equation}

\paragraph{Subnetwork Extraction}
After training, the continuous mask is converted into a deterministic binary mask by thresholding the sigmoid of the learnable logit $l_i$ with Equation~(\ref{eq:binary_mask_thresholding}).

Applying this mask with Equation~(\ref{eq:get_final_weights}) yields a sparse subnetwork, which retains task-relevant connections while removing redundant parameters by assigning them to zero.

\section{Experiments}

\subsection{Task Definition}
Two experimental environments were used to evaluate our method: a simplified Minigrid environment and an underwater navigation simulation in HoloOcean. We utilize a pretrained network in \cite{Melvin2026} and follow its experimental configuration, in which different marine species are represented as colored objects and encoded in the context variables using one-hot encoding.

\subsubsection{Navigation in Minigrid}
\begin{figure}
    \centering
    \includegraphics[
    width=0.95\linewidth,
    ]{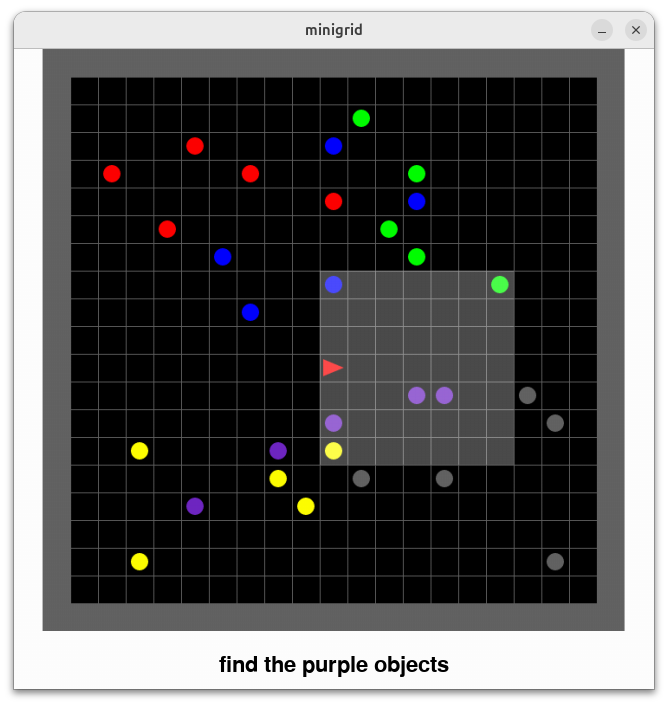}
    \caption{In the Minigrid environment, the agent represented by a red triangle navigates to collect objects of a specified color. The shaded rectangle overlay denotes the agent’s field of view.}
    \label{fig:minigrid_find_many_obj_task}
\end{figure}

A toy navigation task in the Minigrid environment, as shown in Fig.~\ref{fig:minigrid_find_many_obj_task}, was employed to develop and test the method. In this task, the agent, represented as a red triangle, should navigate the environment to collect objects of the specified color. The shaded rectangle overlay in the figure indicates the agent’s field of view, representing its partial observation of the environment. For subnetwork discovery experiments, we used a pretrained MTRL network trained on different tasks jointly for collecting red, blue, purple, and grey objects, with each color represented as a one-hot context variable.

\subsubsection{Underwater Navigation in HoloOcean}
\begin{figure}
    \centering
    \includegraphics[
    width=0.95\linewidth,
    ]{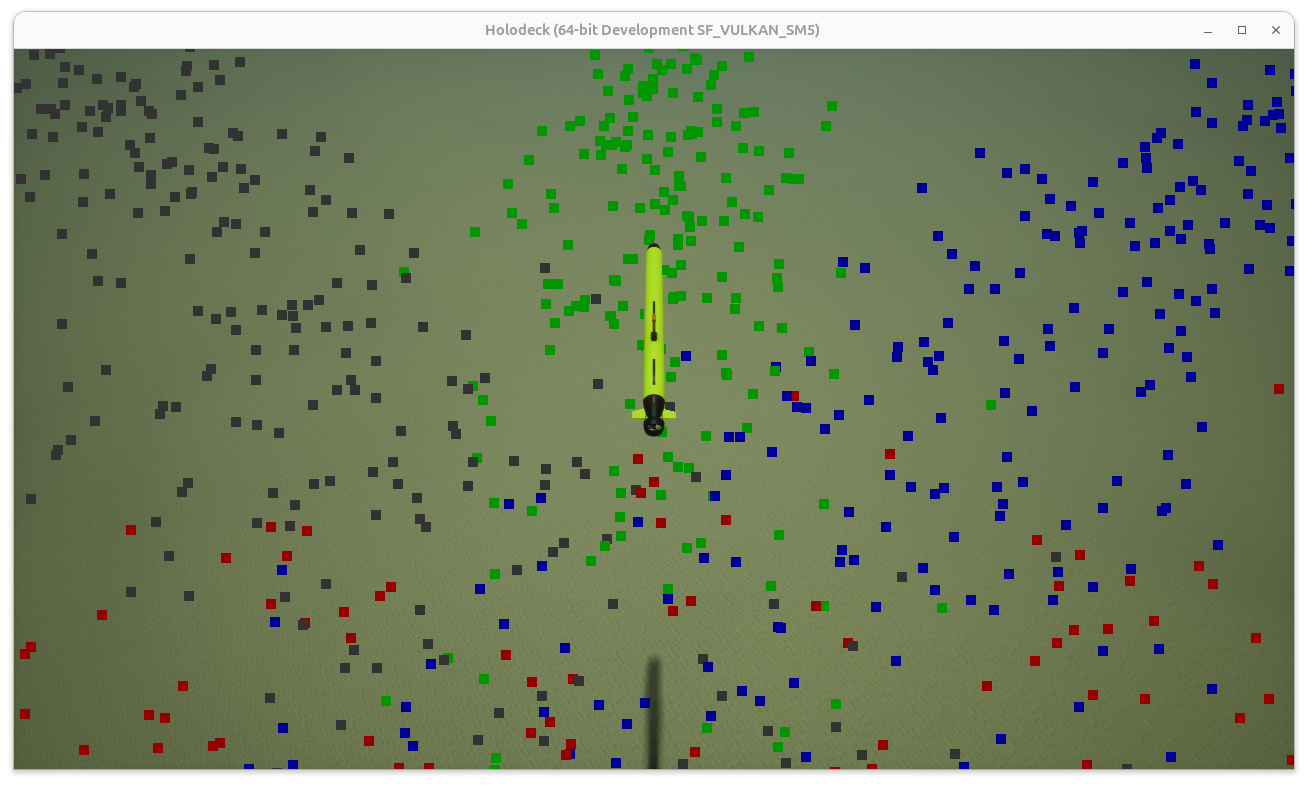}
    \caption{In the HoloOcean environment, the yellow AUV navigates through the environment to pass by specified species, represented as squares of different colors, with each color corresponding to a distinct species.}
    \label{fig:holoocean_env}
\end{figure}

In the underwater navigation tasks simulated in HoloOcean, as illustrated in Fig.~\ref{fig:holoocean}, the agent should navigate the AUV from a starting position to find a specified species, such as crabs, shells, or octopuses. Only finding the species matching the context variable is considered correct. Experiments are conducted using various target species and spatial configurations to analyze how the network leverages shared knowledge across tasks. In our experimental environment, a variety of colored objects are used, as shown in Fig.~\ref{fig:holoocean_env}, with each color representing a distinct species.

\subsection{Experimental Analysis}
Our experiments aim to understand the internal structural of networks in contextual MTRL, focusing on three main aspects: the performance of task-specific subnetworks, the distribution of shared weights across tasks, and the role of context variables. Analyses are conducted separately for the Minigrid and HoloOcean environments.

\subsubsection{Subnetworks Performance}
We identify subnetworks responsible for navigation toward different goals and set all weights not belonging to each selected subnetwork to zero. We then evaluate performance by comparing the normalized RL return of each subnetwork with that of the original full network on the task on which it was trained. The average return is computed as the total reward accumulated across all evaluation episodes, divided by the number of episodes. This quantity is subsequently normalized using the minimum and maximum achievable returns in each environment, respectively.

\subsubsection{Shared Weights across Tasks}
Subnetworks are compared across tasks to identify weights used by multiple tasks, revealing where shared knowledge in MTRL is represented. We visualize partially shared weights, which are activated in only some subnetworks, as well as globally shared weights, which are used across all subnetworks.

\subsubsection{Context Variables} 
The usage of context variables within subnetworks is analyzed, along with subnetwork performance on other tasks, revealing the role of context in task identity.

\subsection{Experimental Results}

\subsubsection{Subnetworks Performance}
A comparison of normalized average returns between the full network and task-specific subnetworks on their corresponding pruning tasks indicates that the extracted subnetworks preserve task-relevant knowledge.

In MiniGrid, the subnetworks generally achieve their highest performance on the tasks for which they were derived, while exhibiting reduced performance on other tasks, consistent with effective task specialization. However, this trend does not hold uniformly across all subnetworks, as the blue subnetwork in particular shows weaker and less consistent performance. In contrast, the subnetworks corresponding to the purple and red tasks exhibit minimal performance degradation on their respective tasks, indicating more stable task-specific representations.

For the navigation task in HoloOcean, the full network exhibits a tendency toward a suboptimal policy with largely uniform performance across tasks. Nevertheless, the corresponding pruned subnetworks retain comparable performance, suggesting that pruning does not degrade the learned behavior and may instead isolate the most relevant task-specific structure.

\begin{figure}[htbp]
    \centering

    \begin{subfigure}{\linewidth}
        \centering
        \includegraphics[
        width=0.98\linewidth,
        trim=0.5cm 0.4cm 0.5cm 1.5cm,
        clip
        ]{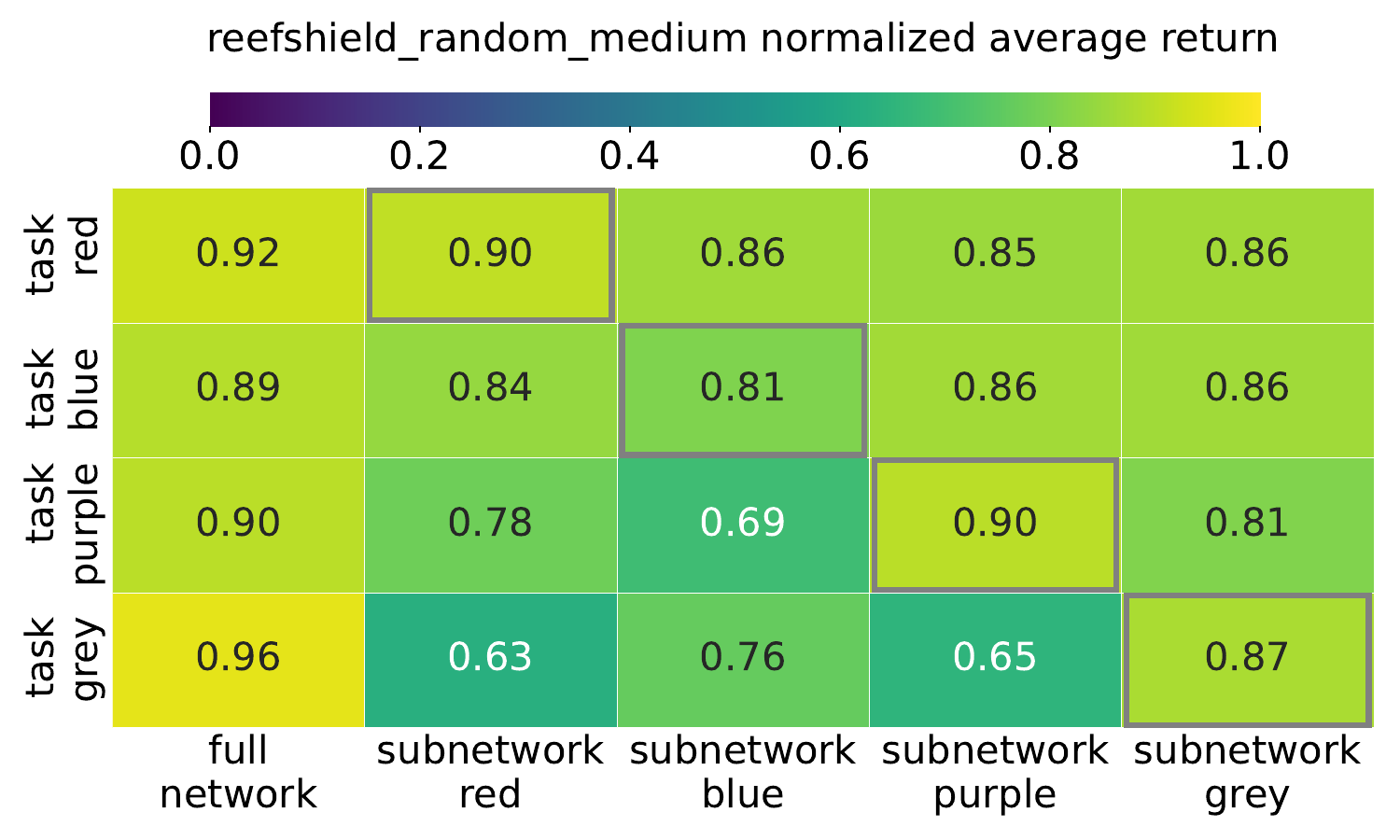}
        \caption{MiniGrid normalized average return}
        \label{fig:minigrid_avg_return_heatmap}
    \end{subfigure}

    \vspace{0.4cm}

    \begin{subfigure}{\linewidth}
        \centering
        \includegraphics[
        width=\linewidth,
        trim=0.5cm 0.4cm 0.5cm 3.5cm,
        clip
        ]{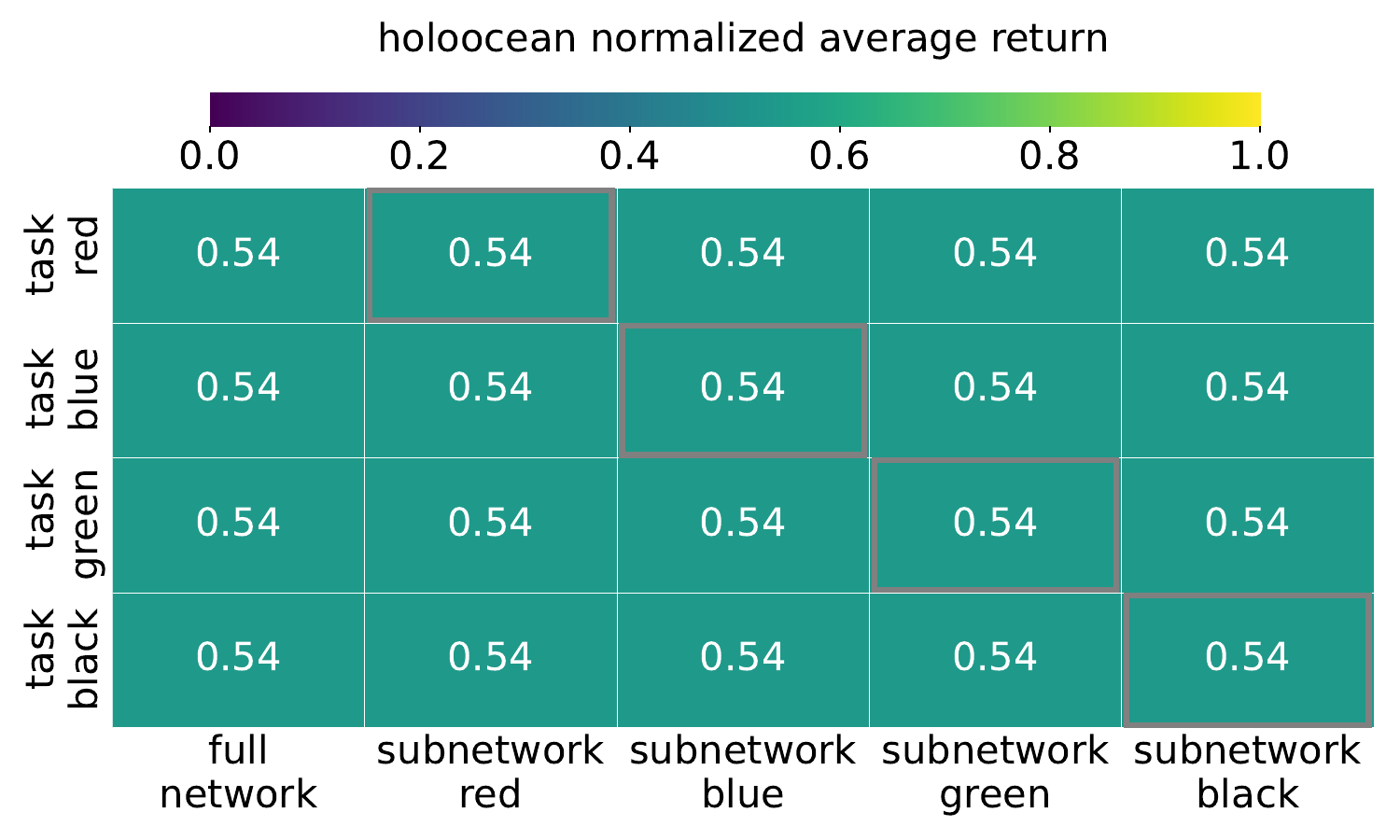}
        \caption{HoloOcean normalized average return}
        \label{fig:holoocean_avg_return_heatmap}
    \end{subfigure}
    \caption{A comparison of normalized average returns between the full network and task-specific subnetworks on their corresponding pruning tasks indicates that the extracted subnetworks preserve task-specific knowledge.}
    \label{fig:avg_return_heatmap}
\end{figure}

Overall, performance on non-corresponding tasks decreases more noticeably than on the corresponding target task after applying the learned masks, in the case of a well-trained full network. Otherwise, when the full network is not well optimized, the pruned subnetworks also do not exhibit a marked performance degradation. The corresponding performance results are shown in Fig.~\ref{fig:avg_return_heatmap}.

\subsection{Shared Weights across Tasks}
\begin{figure}
    \centering
    \includegraphics[
    width=0.98\linewidth,
    trim=2cm 10cm 0cm 13cm,
    clip
    ]{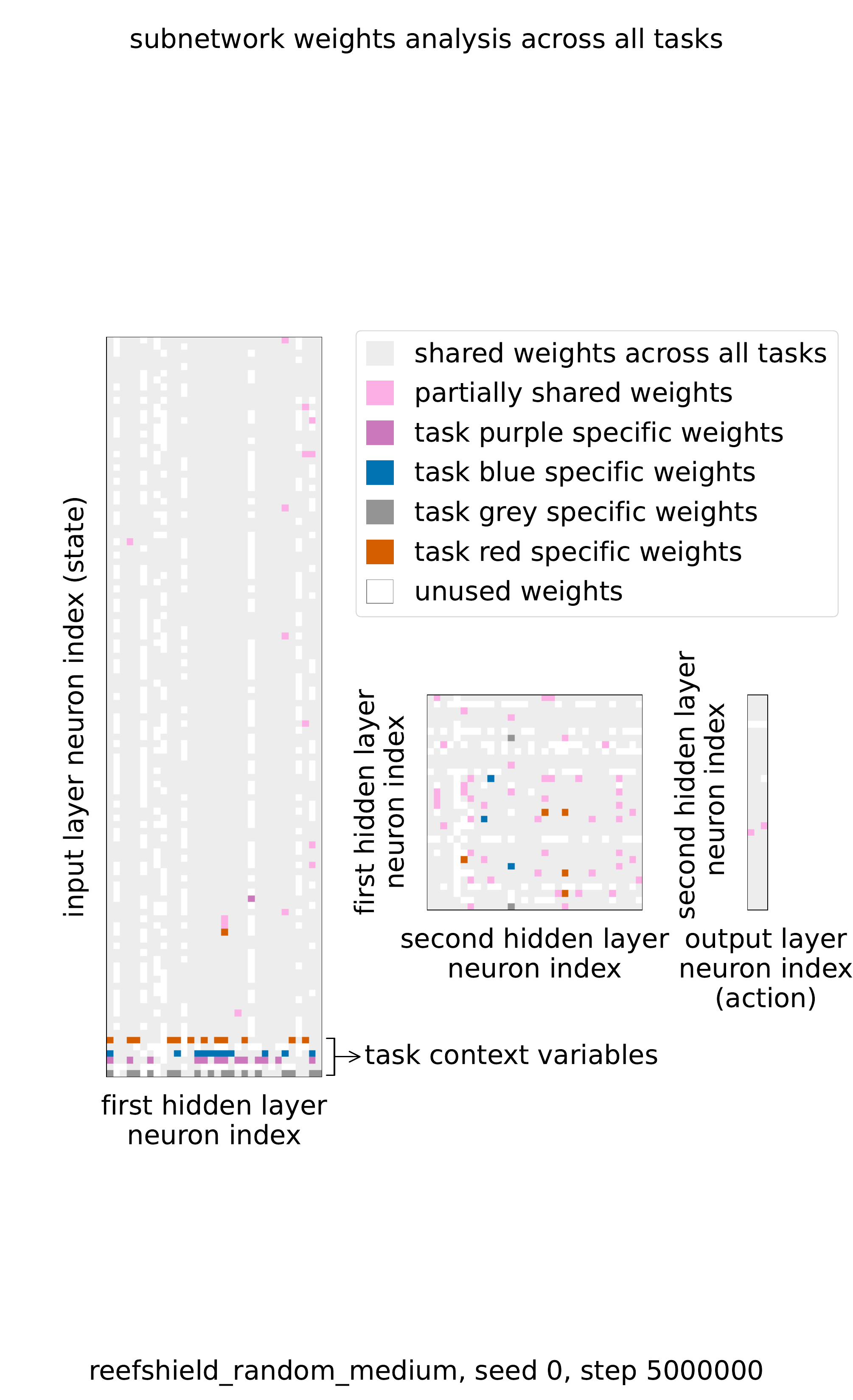}
    \caption{The analysis of shared and task-specific weights across subnetworks of the Minigrid navigation task shows that most parameters are devoted to shared representations, while only a small fraction is task-specific. This suggests that the model learns largely non-redundant common structure across tasks. At the same time, each subnetwork strongly depends on its associated context variable, indicating that contextual information plays a key role in distinguishing tasks and guiding task-specific behavior within the MTRL framework.}
    \label{fig:minigrid_subnet_weights_analysis}
\end{figure}

\begin{figure*}[htbp]
    \centering
    \includegraphics[
    width=\linewidth,
    trim=5cm 3cm 1cm 4cm,
    clip
    ]{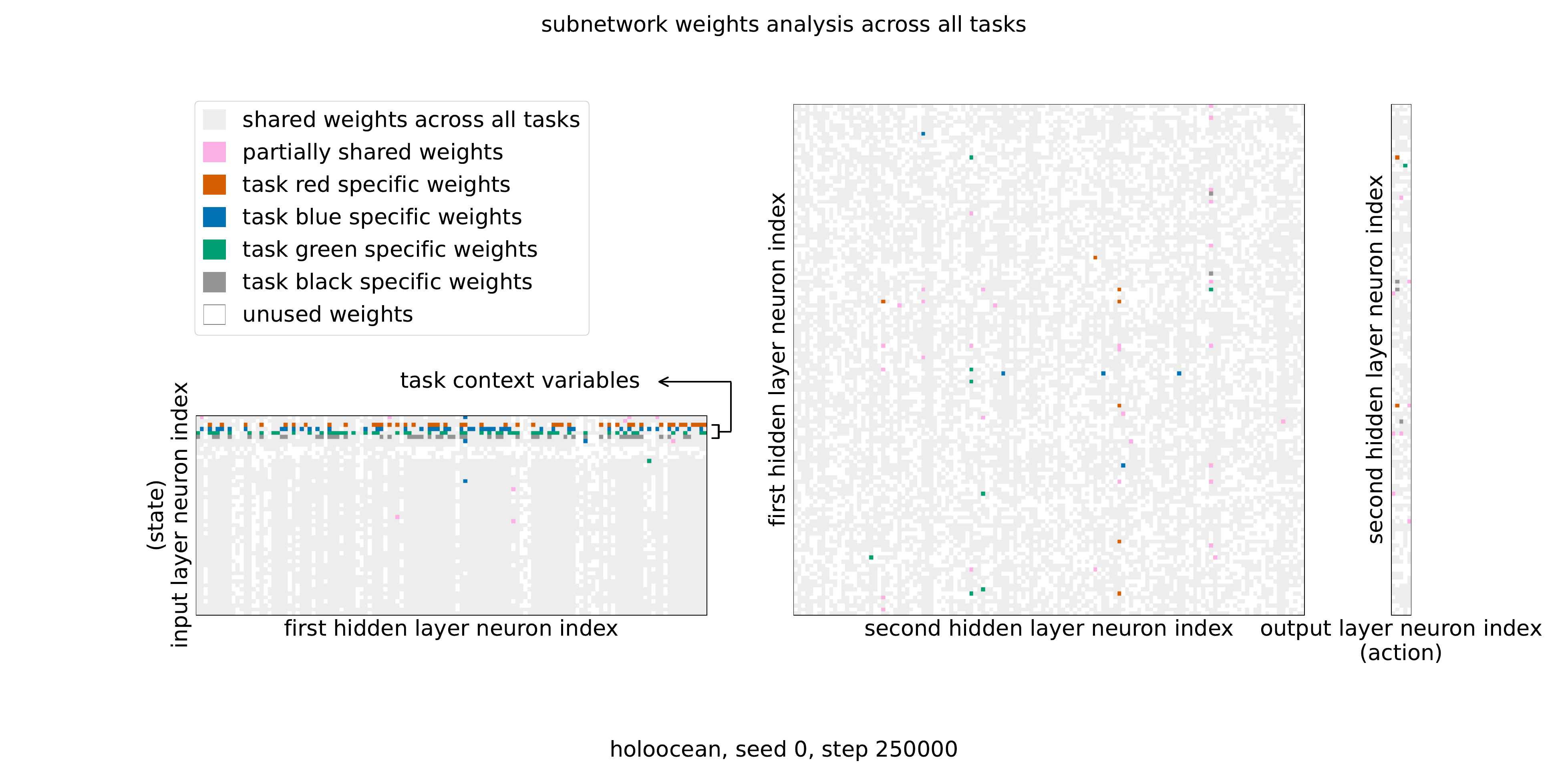}
    \caption{The comparison of weights across subnetworks in the HoloOcean navigation task shows that, after pruning, most parameters are shared, with only a small fraction being task-specific. This suggests that the model learns a largely non-redundant shared representation across tasks, encoded in the shared weights. Moreover, each subnetwork strongly depends on its corresponding task context variable, highlighting the key role of contextual information in distinguishing tasks and guiding behavior within the MTRL framework.}
    \label{fig:holoocean_subnet_weights_analysis}
\end{figure*}

An analysis of shared and task-specific weights across subnetworks for the navigation task in the MiniGrid and HoloOcean environments is presented in Fig.~\ref{fig:minigrid_subnet_weights_analysis} and Fig.~\ref{fig:holoocean_subnet_weights_analysis}, respectively. Following the pruning procedure, approximately 12.82\% and 33.37\% of the weights are removed from the navigation networks in MiniGrid and HoloOcean, respectively. After eliminating inactive weights across all tasks, globally shared parameters constitute approximately 96.84\% and 98.23\% of the total weights in the respective environments, whereas task-specific parameters account for only 1.58\% and 1.45\%.

These results indicate that each task utilizes only a small subset of the overall network capacity, while the majority of parameters are devoted to shared representations. Moreover, this extensive parameter sharing suggests that the learned representations capture largely non-redundant common structure; otherwise, such weights would likely have been pruned. Overall, these findings indicate that the MTRL networks exploit substantial shared knowledge across closely related tasks.

\subsection{Context Variables}
Fig.~\ref{fig:minigrid_subnet_weights_analysis} and Fig.~\ref{fig:holoocean_subnet_weights_analysis} also show that each subnetwork predominantly depends on its associated context variable, underscoring the critical role of contextual information in shaping task-specific behavior. Within the task-specific parameters, approximately 81.25\% and 85.46\% correspond to weights connecting the context variables.

These results further indicate that connections from context variables of trained tasks to hidden-layer neurons are significantly more likely to be classified as task-specific than weights associated with other state inputs. Overall, this suggests that the pretrained network effectively leverages context variables to differentiate between tasks, consistent with the objectives of contextual MTRL.

\section{Conclusion}
The experimental results demonstrate that the pruning process successfully identifies task-relevant subnetworks across most subtasks, with only a slight reduction in task-specific performance. Weight analysis further shows that these subnetworks strongly depend on their associated context variables, highlighting the crucial role of contextual information in guiding task-specific behavior and in distinguishing and exploiting task-relevant information within the contextual MTRL framework. Moreover, a substantial proportion of the retained parameters are shared across tasks, suggesting that the MTRL framework effectively leverages shared representations among related subtasks. This parameter sharing likely contributes to improved sample efficiency and enhanced generalization capability.

In addition, it is observed that several rows and columns of the weight matrices are substantially pruned, indicating that the corresponding neurons contribute minimally to task performance and are therefore less relevant. This finding motivates further investigation into neuron-level pruning and its relationship to task specialization.

Future work may extend this analysis by performing explicit neuron pruning and conducting an ablation study in which weights connecting context variables in the original full network are randomly removed to further validate their functional role. Such an investigation would enable a systematic assessment of the influence and importance of these connections on task-specific performance, thereby providing deeper insights into the role of context variables in shaping behavior within the contextual MTRL framework.

\section*{Acknowledgments}
This work was funded by the German Federal Ministry for the Environment, Climate Action, Nature Conversation and Nuclear Safety (BMUKN) supported by the ZUG under grants 67KIA4036C and 67KIA4036A, and partially supported by the German Federal Ministry of Research, Technology and Space (BMFTR) under the Robotics Institute Germany (RIG) under grant 16ME1010.
The authors would like to thank Alexander Fabisch, Yuhan Jin, and Nayari Lessa for their valuable feedback and discussion on this manuscript. 
\printbibliography

\end{document}